\documentclass[letterpaper, 10 pt, conference]{ieeeconf}  
\IEEEoverridecommandlockouts                              
\usepackage{graphics}    
\usepackage{times}       
\usepackage{amsmath}     
\usepackage{amssymb}     
\usepackage{graphicx}
\usepackage{algorithm}
\usepackage[noend]{algpseudocode}
\usepackage{blindtext}
\usepackage{url}

\usepackage[font=small]{caption} 
\usepackage{tabularx}
\usepackage{makecell}
\usepackage{multirow}
\usepackage{hhline}
\usepackage{subfig}
\usepackage{graphicx}
\usepackage{siunitx}

\usepackage[backend=biber,style=ieee,mincitenames=1,maxnames=20,maxcitenames=1,natbib=true]{biblatex}
\usepackage{doi}
\usepackage{booktabs}
\usepackage[capitalise]{cleveref}
\usepackage[font=small]{caption}

\def\figref#1{Fig.~\ref{#1}}
\def\tabref#1{Tab.~\ref{#1}}
\def\eqref#1{Eq.~(\ref{#1})}

\usepackage{adjustbox}

\title{\LARGE \bf Learning Goal-Directed Object Pushing in Cluttered Scenes \\With Location-Based Attention}

\author{Nils Dengler$^{1,4,5*}$ \and Juan Del Aguila Ferrandis$^{2*}$ \and Jo\~ao Moura$^{2,3}$ \and Sethu Vijayakumar$^{2,3}$ \and Maren Bennewitz$^{1,4,5}$
\thanks{$\ast$ These authors contributed equally to this work.}
\thanks{$^1$: Humanoid Robots Lab, University of Bonn, Germany}
\thanks{$^2$: School of Informatics, The University of Edinburgh, Edinburgh, UK}
\thanks{$^3$: The Alan Turing Institute, London, UK}
\thanks{$^4$: The Lamarr Institute, Bonn, Germany}
\thanks{$^5$: The Center for Robotics, University of Bonn, Germany}
\thanks{This work has partly been supported by the European Commission under grant agreement number 964854 (RePAIR), by the BMBF within the Robotics Institute Germany, grant No. 16ME0999, and by the JST Moonshot R\&D (Grant No. JP-
MJMS2031). 
  }%
}
\begin{document}
\maketitle
\thispagestyle{empty} 
\pagestyle{empty}

\begin{abstract}
In complex scenarios where typical pick-and-place techniques are insufficient, often non-prehensile manipulation can ensure that a robot is able to fulfill its task.
However, non-prehensile manipulation is challenging due to its underactuated nature with hybrid-dynamics, where a robot needs to reason about an object's long-term behavior and contact-switching, while being robust to contact uncertainty.
The presence of clutter in the workspace further complicates this task, introducing the need to include more advanced spatial analysis to avoid unwanted collisions.
Building upon prior work on reinforcement learning with multimodal categorical exploration for planar pushing, we propose to incorporate location-based attention to enable robust manipulation in cluttered scenes.
Unlike previous approaches addressing this obstacle avoiding pushing task, our framework requires no predefined global paths and considers the desired target orientation of the manipulated object.
Experimental results in simulation as well as with a real KUKA iiwa robot arm demonstrate that our learned policy manipulates objects  successfully while avoiding collisions through complex obstacle configurations, including dynamic obstacles, to reach the desired target pose.
\end{abstract}

\section{Introduction} \label{sec:intro}

Incorporating non-prehensile manipulation into a robot's skill set enhances its versatility beyond pick-and-place techniques~\cite{efendi2024technological, stuber2020let}. 
More broadly, non-prehensile manipulation refers to moving or controlling objects without grasping, utilizing techniques such as pushing, rolling, or sliding.
This capability allows robots to manipulate a wider range of ungraspable objects and access to otherwise unreachable grasping configurations through their repositioning and reorientation~\cite{zhou2023learning}.

\begin{figure}[!t]
	\centering
	\includegraphics[width=0.95\linewidth, trim=70 50 0 0, clip]{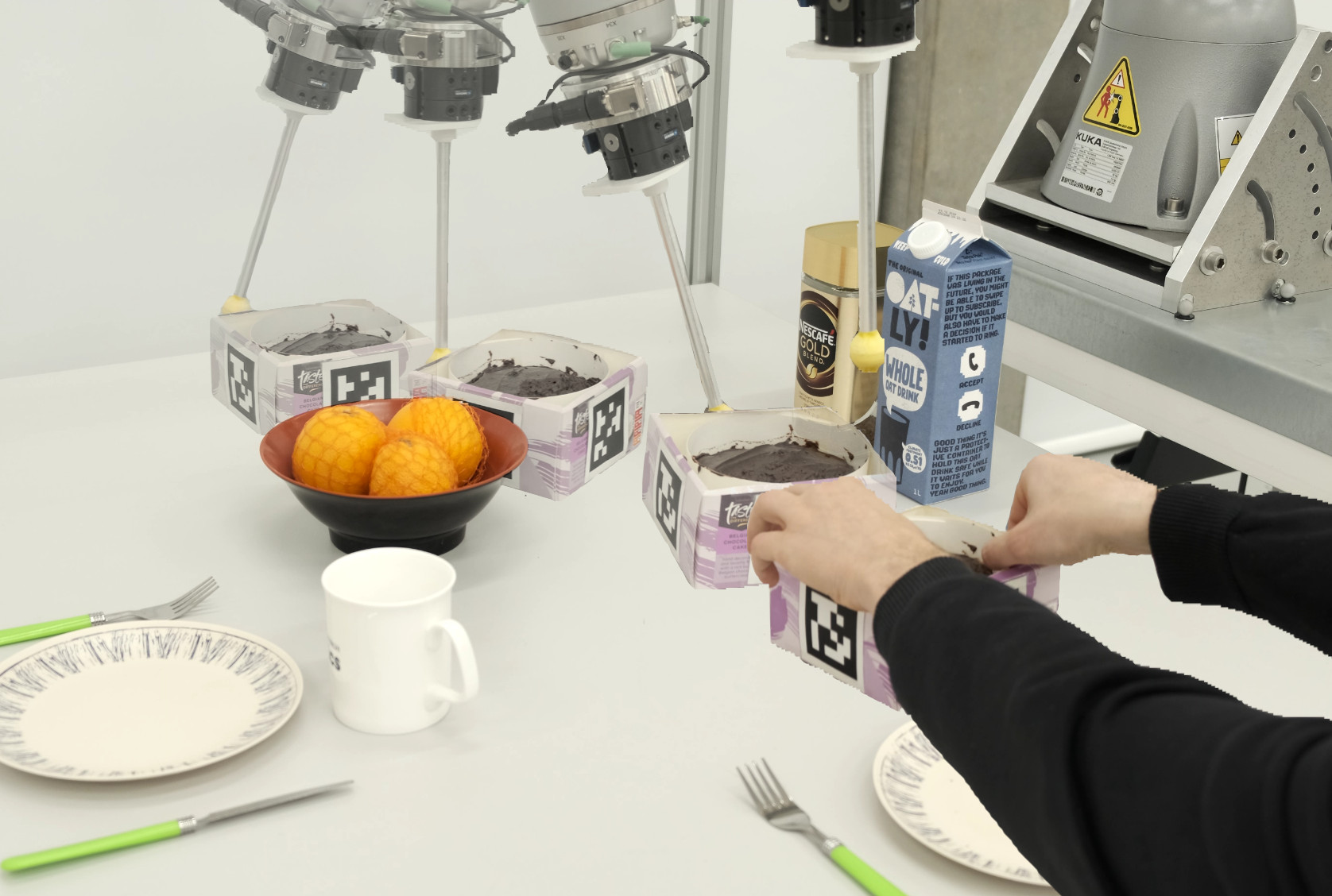}
	\caption{Example scenario for pushing in a cluttered workspace. The robot moves a cake to a specified target pose while avoiding collisions with other objects on the table.
	} 
	\label{fig:cover}
\end{figure}

In cluttered environments, avoiding obstacles introduces a new dimension of complexity to non-prehensile manipulation, requiring advanced long-horizon spatial reasoning that integrates collision constraints while maintaining responsiveness to dynamic and unpredictable elements~\cite{moura2022non}.
Therefore, a real-time scene understanding is essential to predict interactions, generate feasible trajectories, and adapt to both static and dynamic components in the scene.
For example, \cref{fig:cover} shows a scenario in which the robot pushes a cake to a person in order for them to reach it, while avoiding the other items on the table. 

Current research predominantly focuses on precise object pushing in free space \cite{wang2023learning, ferrandis2024learning} or on cluttered surfaces without restricting interactions between the objects~\mbox{\cite{wu2023efficient, bejjani2021learning}}.
Only few studies consider pushing in cluttered environments while incorporating \mbox{collision constraints~\cite{dengler2022learning, pasricha2022pokerrt}}. 
However, they rely on pre-computed path guidance and scale poorly to more complex scenarios~\cite{leve2024explicit}. 
Recently,~\citet{ferrandis2023iros} demonstrated significant performance improvements in free-space pushing tasks by leveraging model-free reinforcement learning (RL) with categorical exploration to capture the multimodal behavior arising from the different possible contact interaction modes between the robot and the manipulated object.

Therefore, we propose a system for pushing in cluttered workspaces that builds upon \cite{ferrandis2023iros} but incorporates an occupancy grid map state representation to capture the clutter layout.
In contrast to prior RL work~\cite{dengler2022learning}, we avoid relying on precomputed guidance, such as a global path, as it can restrict the RL agent in its exploration process.
Predefined paths limit the flexibility of the RL agent, preventing it from discovering alternative, potentially more efficient strategies for pushing in cluttered environments.
For example, a slightly longer path that avoids dense clutter might be preferable over a shorter but more obstructed path.
Additionally, by using a more general representation, i.e., an occupancy grid map, our agent generalizes to unseen scenarios, such as dynamic or differently shaped objects, compared to fixed representations with specific object information.

However, high-dimensional representations incur higher computational costs and make learning more complex, as they increase the number of parameters and the dimensionality of the search space, which is particularly problematic when learning online with RL. 
To address this, we investigate the use of a lightweight attention mechanism, called location-based attention \cite{luong2015effective}, to extract and selectively focus on relevant spatial features from the environment state. 
In our experiments, we demonstrate successful goal-oriented pushing behavior, combining categorical exploration with attention-based feature extraction to effectively handle cluttered environments.

To summarize, the key contributions of our work are:
\begin{itemize}
    \item A guidance-free RL framework for obstacle-avoiding non-prehensile object pushing in cluttered environments that eliminates the need for global path planning. 
    \item The integration of a lightweight, location-based attention mechanism for spatial reasoning from occupancy grids, allowing the RL agent to attend to task-relevant features while maintaining computational efficiency and generalization to unseen obstacle configurations.
    \item Extensive experimental validation, including: \textbf{(i)} Simulated evaluations across a variety of unseen obstacle shapes and layouts, demonstrating the robustness and adaptability of the learned policy; \textbf{(ii)} Ablation studies showing the performance benefits of location-based attention over common alternative feature extractors; \textbf{(iii)} 
    Real-world deployments on a KUKA iiwa robot in diverse and dynamic scenarios, confirming the policy’s robustness, transferability, and smooth, continuous trajectory execution without reliance on offline path planning.
\end{itemize}

\section{Related Work}

\label{sec:related}

Previous works developing model-based robot controllers for planar pushing generally use Model Predictive Control~(MPC) to track nominal trajectories computed offline~\cite{wang2024uno, moura2022non, baumeister2024incremental}.
These approaches achieve smooth and highly precise pushing motions. 
However, due to the short-horizon of MPC, 
large disturbances to the manipulated object or significant changes in the obstacle layout require offline re-computation of the nominal trajectory.
We overcome this problem by using an RL agent, trained on different scenarios, that dynamically adapts its policy in real-time based on changes in the environment.

Other approaches apply model-free methods, primarily RL.
Many of these works focus on learning pushing policies for clutter-free environments~\cite{bergmann2024precision,ferrandis2023iros,wang2023learning}.
Another prominent research direction is the synergy of pushing and grasping actions to retrieve objects from clutter~\mbox{\cite{wu2023efficient, liu2023synergistic, jiang2023contact}}.
However, the characteristics here are different from the task we consider, since their goal is to move the clutter away to reach and retrieve the target object through a grasping action, hence disregarding collision constraints.

Only few studies consider pushing in cluttered
environments while incorporating collision constraints~\mbox{\cite{krivic2019pushing, pasricha2022pokerrt, dengler2022learning}}.
In particular,~\citet{pasricha2022pokerrt} use Rapidly-exploring Random Trees~(RRT) to poke an object while avoiding obstacles in the workspace. 
This method results in non-smooth motions that are unable to accurately control the resulting object pose. 
Furthermore, RRT scales poorly for non-prehensile pushing tasks~\cite{leve2024explicit}.
\citet{krivic2019pushing} utilize a precomputed corridor to constrain the object and robot within defined boundaries during pushing. However, in narrow scenes, this approach is prone to local minima, often resulting in oscillatory robot behavior.

To the best of our knowledge, the work proposed by~\citet{dengler2022learning} is the only other model-free learning-based approach that addresses the problem considered in this paper.
However, their approach relies on various assumptions that reduce the complexity of the problem.
Most significantly, they use sub-goals from a pre-computed global path in order to guide the policy towards the target position. 
Furthermore, the authors consider only a 2D~target position, neglecting the orientation of the object.
In contrast, we present a guidance-free method that avoids the drawbacks of using pre-computed global paths and considers both the target position and orientation of the manipulated object. 

For feature extraction, attention-based approaches have recently gained significant popularity \cite{hassanin2024visual,yuan2024transformer,Manchin_2019}, e.g., in navigation tasks \cite{dawood2023safe,cao2023ariadne}, due to their ability to extract relevant features from the input while maintaining low computational cost, which is crucial for training RL policies with highly parallelized environments.
One subclass of these algorithms is location-based attention \cite{luong2015effective, niu2021review}, which assigns attention weights to selectively focus on input features based on their spatial location without having to compute relationships between all pairs of the input data.
This feature significantly reduces the computational complexity, particularly in high-dimensional input spaces such as the occupancy grid maps we use in this work, where traditional attention mechanisms, like multi-head self-attention, can be computationally expensive due to the large number of pairwise relationships they calculate.
Recently,~\citet{deheuvel2023spatiotemporal} leveraged location-based attention within an RL approach for robot navigation among obstacles.
Their method still relies on sub-goals sampled from a global path, which we aim to overcome. 
In this paper, we show that explicit global guidance is unnecessary, as the attention module can extract sufficient features from the occupancy grid representation to achieve goal-directed and obstacle-avoiding pushing behavior.

\begin{figure*}[t]
	\centering
	\includegraphics[width=0.8\textwidth, trim=35 85 35 85, clip]{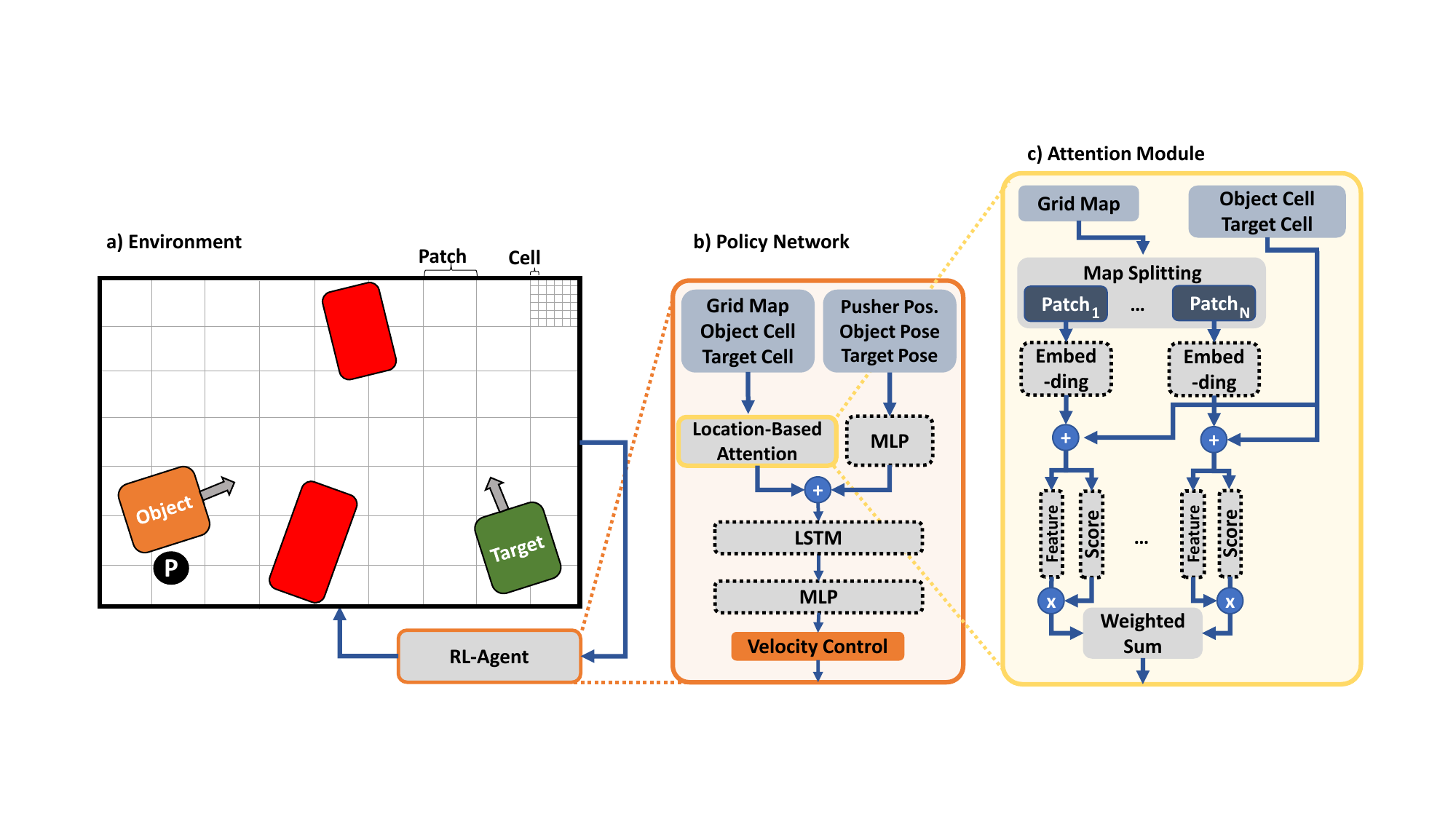}
	\caption{Overview of our framework for learning goal-directed pushing using location-based attention. 
    (a) The grid map of the environment together with the object and target pose, as well as the position of the pusher is fed to the RL-agent (b). In comparison to previous work~\cite{dengler2022learning}, we use a location-based attention module (c) for feature extraction of the cluttered scene. 
	} 
	\label{fig:architecture}
    \vspace{-5px}
\end{figure*}

\section{Method}
\label{sec:main}
In this work, we consider the following problem. 
A robotic arm aims to push an object from its current pose to a target pose~$(x, y, \theta)$ within a bounded planar workspace with its end effector, i.e., the pusher. 
In addition to the pushed object, there are other objects in the workspace which are obstacles the pushed object needs to avoid. 

To address this problem, we propose an RL framework that leverages categorical exploration \cite{ferrandis2023iros} to capture the multimodal nature of planar pushing, as well as location-based attention to extract and selectively focus on relevant spatial features from the workspace occupancy grid, achieving obstacle avoidance while manipulating the object towards the target pose.
In the following, we describe the design of our RL framework, summarized in Fig.~\ref{fig:architecture}.

\subsection{Feature Extraction}
\label{sec:feature}
\vspace{-5px}
\subsubsection{\textbf{Preprocessing}}
At the beginning of each episode, we generate a binary occupancy grid of the workspace, where $1$ represents obstacle and $0$ free space.
We use a resolution of $\SI{0.005}{\meter} \times \SI{0.005}{\meter}$ per grid cell.

\subsubsection{\textbf{Location-Based Attention}}
\label{sec:attention}

Drawing inspiration from Visual Transformers \cite{dosovitskiy2020image}, we decompose the occupancy map into $n$ patches, each of size $P_s=16 \times 16$, where $n \cdot P_s$  matches the size of the original map.
We use a multilayer perceptron (MLP) of size $(192, 128)$ to embed each patch, as depicted in \ref{fig:architecture}.a, encoding its features. 
This encoding process allows us to capture the essential characteristics of each patch, including obstacles and potential paths. 

To provide positional context for each patch in the current task configuration, we concatenate them with the object and target positions, relative to the upper-left corner of each patch.
From the patch embeddings and the positional context, we obtain the attention features and scores using separate MLPs of size $(128, 100, 64)$.
Finally, we compute the weighted attention features as depicted in Fig. \ref{fig:architecture}.c and feed the output of the location-based attention module to the RL agent.

\subsection{Reinforcement Learning}
\label{sec:RL}

The hybrid dynamics inherent in non-prehensile planar manipulation, characterized by varying contact modes such as sticking, sliding, and separation \cite{hogan2020reactive}, make traditional unimodal exploration strategies, generally parametrized through multivariate Gaussian distributions, suboptimal. 
These strategies struggle to model the multimodal nature of interactions that arise from discrete contact transitions. 
Building on recent work in RL for accurate planar pushing~\cite{ferrandis2023iros}, we adopt the on-policy RL algorithm Proximal Policy Optimization~(PPO)~\cite{schulman2017proximal}, using a discretized action space to enable multimodal categorical exploration.

Below, we detail the main components of the RL pipeline.
\subsubsection{\textbf{Observation}}
\label{sec:Observation}
The policy observation of the environment consists of the object and target poses $(x,y,\theta)$, the pusher position $(x,y)$, and the binary occupancy grid that encodes the clutter layout.
To reduce the computational cost during training, we keep the grid layout fixed throughout each episode.
Nevertheless, we show in our hardware experiments that the grid representation can be updated in real time using, e.g., point cloud data or motion capture, and that the learned policies are robust to dynamic changes in the obstacle layout.

\subsubsection{\textbf{Action}}
\label{sec:Action}
We define the policy action as $(v_x, v_y)$, the $x$ and $y$ velocity of the pusher.
Furthermore, we limit the velocity on each axis to the range $[-0.1, 0.1]$ \si{\meter\per\second} and use $0.02$ \si{\meter\per\second} velocity steps for each categorical bin.

\subsubsection{\textbf{Reward}}
\label{sec:Reward}
We define our reward function $r_{total}$ as
\begin{equation}
    r_{total} = r_{term} + k_1 (1 - r_{dist}) + k_2 (1 - r_{ang}) + r_{coll},
\end{equation}
with $k_1, k_2$ being scaling factors.
$r_{term}$ is a large sparse termination reward, which is positive when the object reaches the desired target pose and otherwise negative.
$r_{dist}$ is the Euclidean distance of the manipulated object to the target position, normalized to the range $[0,1]$, and $r_{ang}$ the angular distance of the object to the target orientation, also normalized to $[0,1]$.
In addition, we use $r_{coll}$ as a binary negative reward to penalize at every step any kind of contact with an obstacle by the pusher or the object. 
If there is no collision during one time step then $r_{coll}=0$.
\subsubsection{\textbf{Policy and Value Networks}}
We use the same architecture for the policy and value networks~(see \cref{fig:architecture}.b).
In particular, the attention module extracts weighted attention features (size $64$) from the occupancy grid.
We also use an MLP (size $64$) to extract features from the remaining observation, which consists of the object and target pose, as well as the pusher position.
We concatenate these two feature vectors and feed them through a Long Short-Term Memory (LSTM) (size $256$) layer and an MLP (size $128$) layer. 
Using LSTMs for the policy and value networks enables to capture the hidden temporal dynamics of the environment, including friction and inertia.
The final output of the value network is of size $1$, corresponding to the state value estimate, while the policy network returns a vector of size~$22$, corresponding to logits that define the two categorical distributions for the velocities on the $x$ and $y$ axes. 
For an extensive experimental evaluation of the effectiveness of categorical action distributions and LSTM-based temporal modeling over standard MLP architectures for the uncluttered planar pushing task refer to \cite{ferrandis2023iros}. We did not observe any deviation from these results in our obstacle avoidance task.

\section{Experimental Results}
\label{sec:exp}

\begin{table}[t]
    \centering
    \resizebox{\linewidth}{!}{
    \begin{tabular}{l c | l c}
        \multicolumn{2}{c|}{\textbf{Hyperparameter Values}} & \multicolumn{2}{c}{\textbf{Sampling Distributions}} \\\rule{0pt}{4pt}
        \textbf{Parameter} & \textbf{Value} & \textbf{Parameter} & \textbf{Distribution} \\
        \hline
        Grid Size & $100 \times 140$ & Static Friction & $\mathcal{U}[0.5,0.7]$ \\
        Parallel Environments & $1,440$ & Dynamic Friction & $\mathcal{U}[0.2,0.4]$ \\
        Batch Size & $14,400$ & Restitution & $\mathcal{U}[0.4,0.6]$ \\
        Rollout Length & $120$ & Object Mass & $\mathcal{U}[0.4,0.6]$ \SI{}{\kg} \\
        Update Epochs & $5$ & Object Scale & $\mathcal{U}[0.9,1.1]$ \\
        Clip range $(\epsilon)$ & 0.2 & Obstacle Scale & $\mathcal{U}[0.8,1.2]$ \\
        Discount factor $(\lambda)$ & 0.99 & Pusher Scale & $\mathcal{U}[0.95,1.05]$ \\
        GAE parameter $(\gamma)$ & 0.95 & Position Noise & $\mathcal{N}[0, 0.001^2]$ \SI{}{\meter} \\
        Entropy bonus coefficient & 0 & Orientation Noise & $\mathcal{N}[0, 0.02^2]$ \SI{}{\radian} \\
        Value function coefficient & 0.5 & & \\
        Optimizer & Adam & & \\ 
    \end{tabular}
    }
    \caption{Hyperparameter values for RL training and sampling distributions for dynamics randomization and observation noise. $\mathcal{U}$~denotes the uniform distribution and $\mathcal{N}$ the normal distribution.}
    \label{tab:parameters}
\end{table}

\begin{figure}[!t]
	\centering
	\includegraphics[width=.85\linewidth,  trim={0 0 0 0},clip]{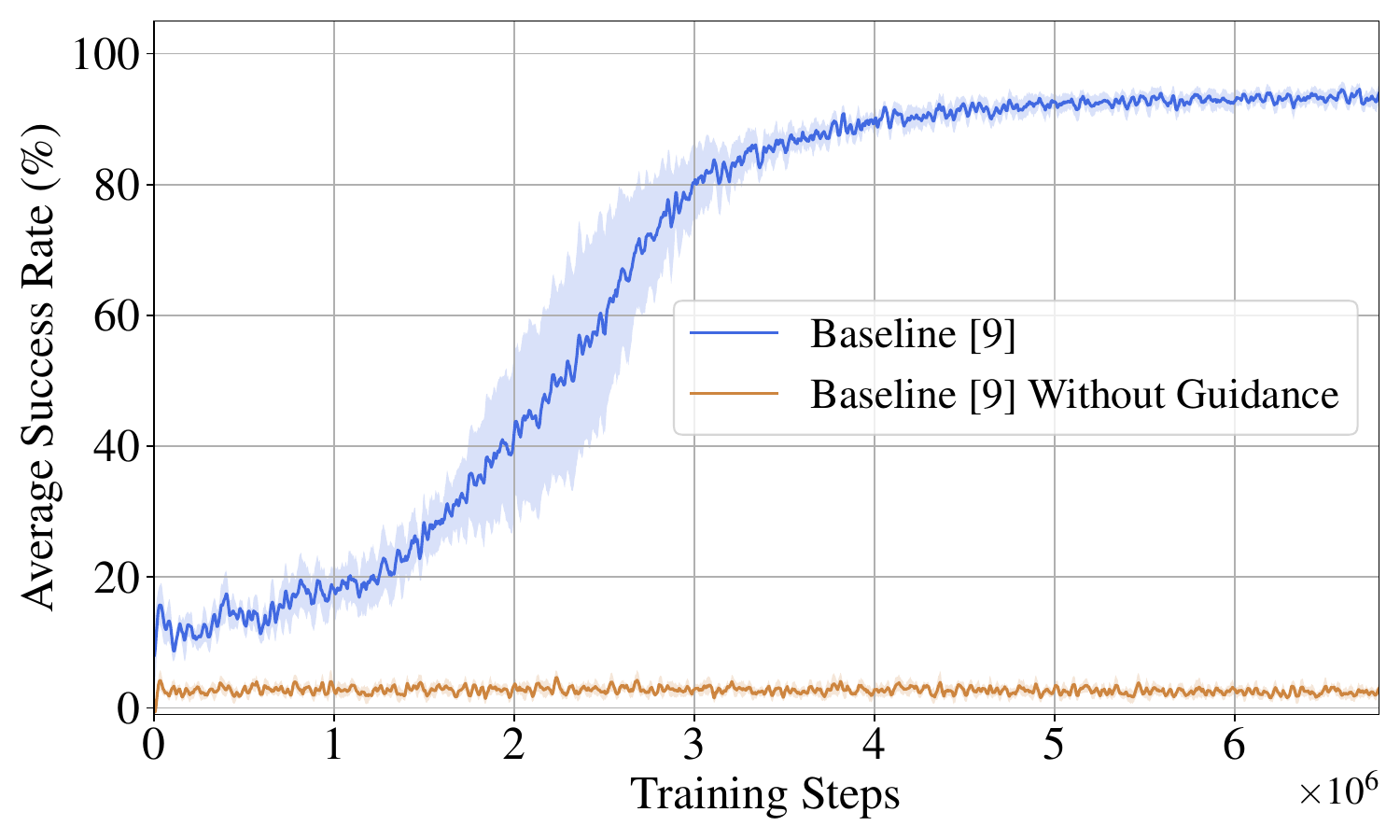}
	\caption{Training performance of the baseline approach \cite{dengler2022learning} (blue), as well as a variant without global path guidance (orange).} 
	\label{fig:Baseline}
    \vspace{-5px}
\end{figure} 

\begin{figure*}[t] \centering 
\subfloat[Training setup\label{eval_environments:a}]{\includegraphics[width=.3\textwidth, trim={1cm 0cm 3cm 4cm},clip]{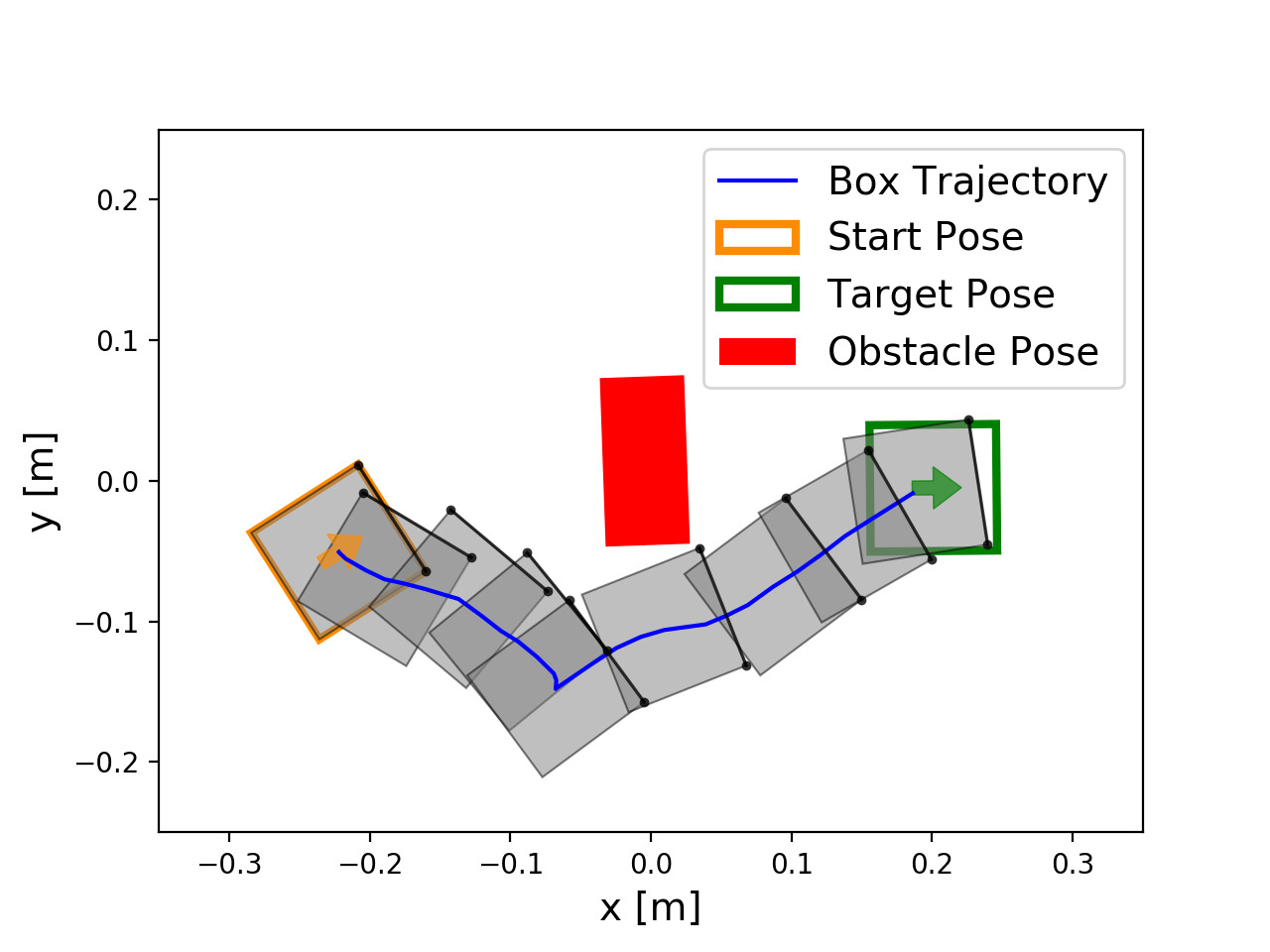}} 
\subfloat[L-shape obstacle\label{eval_environments:b}]{\includegraphics[width=.3\textwidth, trim={1cm 0cm 3cm 4cm},clip]{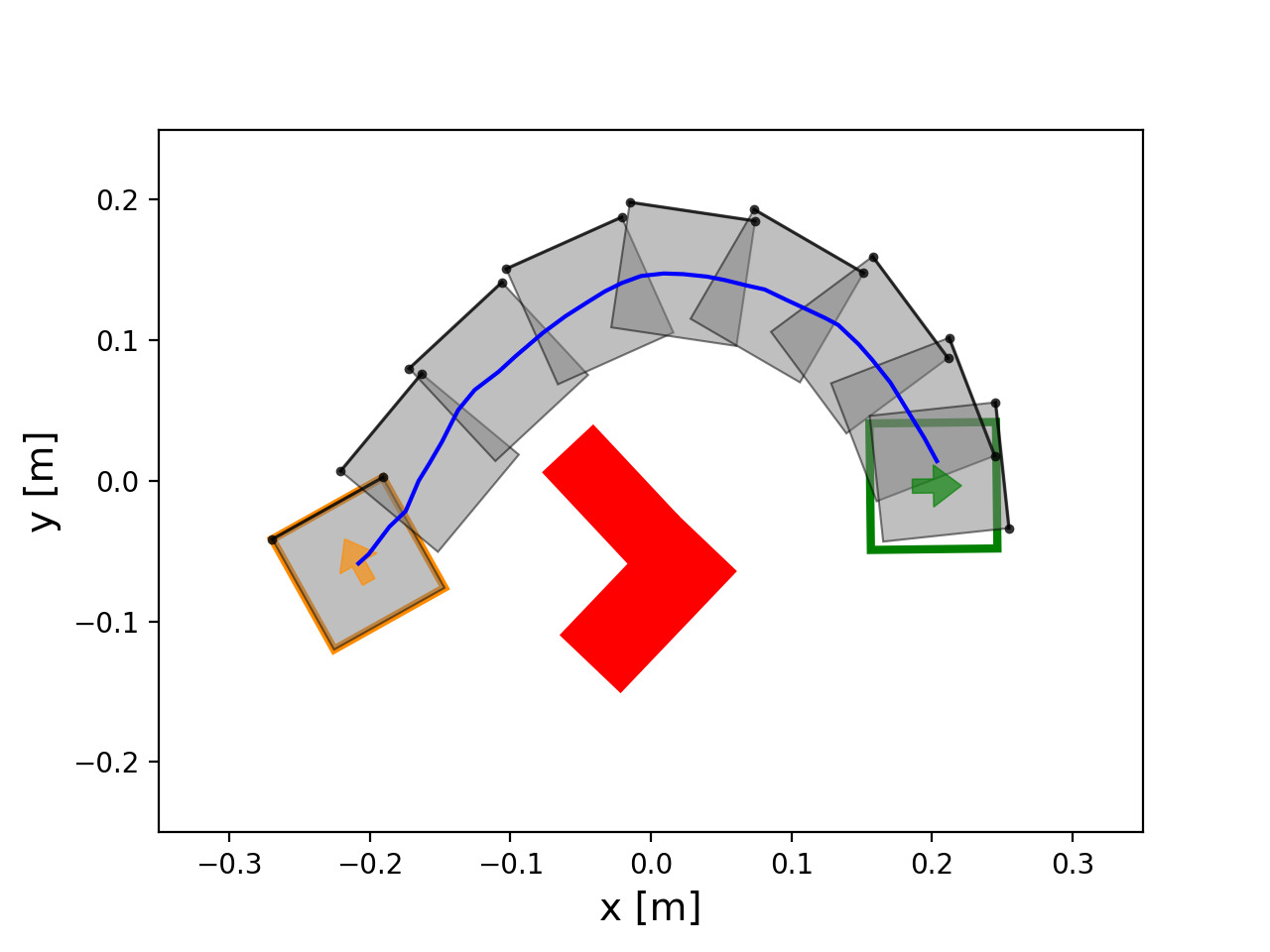}} 
\subfloat[Dual obstacle\label{eval_environments:c}]{\includegraphics[width=.3\textwidth, trim={1cm 0cm 3cm 4cm},clip]{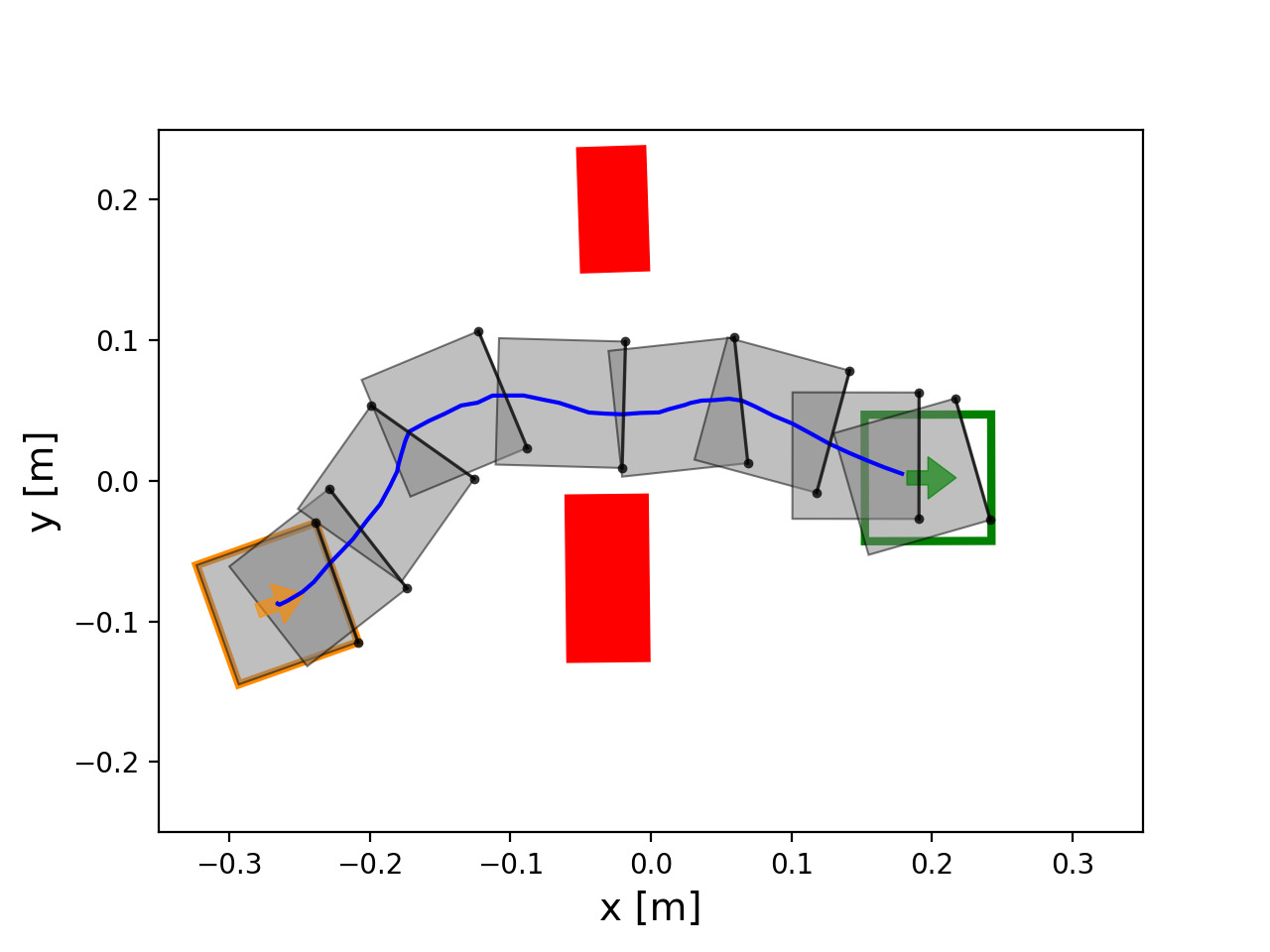}} 
\vspace{-5px}
\caption{Different obstacle configurations and the corresponding trajectories resulting from executing the push actions generated by the RL policy with location-based attention in the physical hardware setup. 
The three experiments show (a) pushing behavior with contact surface switching, (b) a smooth trajectory around an L-shaped obstacle, and (c) a precise pushing maneuver to fit the object through a narrow gap between two obstacles.} 
\label{eval_environments}
\end{figure*} 
\subsection{Model Training}
We train the agents using the Isaac Sim physics simulator~\cite{mittal2023orbit}, developing a custom environment for pushing in clutter to leverage the advantages of massively parallel RL environments.
To accelerate the simulation and RL training, we abstract the robotic model as a spherical pusher and use a single rectangular obstacle as the standard training setup, while additionally fine-tuning with two-obstacle scenarios.
At the start of each episode, we sample random poses for the pusher, the object, the obstacle, and the target, such that the obstacle is between the object and the target.

The policies run at a frequency of \SI{10}{\hertz} and, during training, we enforce a maximum episode length of $160$ steps. 
During evaluation, since we consider more complex scenarios, such as unseen obstacle shapes and multiple obstacles, we increase the maximum episode length to $200$ steps. 
For the reward function, we use a termination reward $r_{term} = 50$, when the episode is successful, and $r_{term} = -10$ when it is unsuccessful due to a violation of workspace boundaries. 
Furthermore, the collision penalty is $r_{coll} = -5$, and we use scaling factors $k_1 = 0.1$,  $k_2 = 0.02$ for the position and angular distance reward terms.  

We use the PPO algorithm with the hyperparameter values specified in the left part of \tabref{tab:parameters}.
Note that we use an adaptive learning rate schedule based on the KL divergence of the policy network~\cite{rudin2022learning} with a target KL divergence of $0.01$. 
Furthermore, if an episode terminates upon reaching the maximum length, we bootstrap the final reward using the state value estimate from the value network~\cite{pardo2018time}.

To bridge the sim-to-real gap, we use dynamics randomization and synthetic observation noise during policy training. 
The right part of \cref{tab:parameters} shows the randomized parameters and corresponding sampling distributions.
We generate correlated noise, sampled at the beginning of every episode, as well as uncorrelated noise, sampled at every step, and add it to the policy observation of the object pose and the pusher position. 
The implementation of our method is publicly available on Github\footnote{\url{https://github.com/LearnToPush/attention-pushing}}.

\subsection{Baseline and Influence of Path Guidance}
\label{sec:baseline}

Since the work of~\citet{dengler2022learning} is the most closely related to our task, we re-implement their approach using PyBullet \cite{coumans2021} and apply it to our obstacle avoidance pushing task. 
We choose PyBullet because their method is unsuitable for GPU parallelization, due to their need for precomputed global paths, making integration with Isaac Sim problematic.
For this analysis, we disregard the orientation of the target object, following~\cite{dengler2022learning}. 
We initially attempted to incorporate the target orientation by including it in both the policy observation and reward function as in our method; however, it led to convergence failure.
Additionally, unlike in \cite{dengler2022learning}, we validate our approach on the physical robotic hardware and, hence, our method includes dynamics randomization and synthetic observation noise.

We train a baseline using our re-implementation of \citet{dengler2022learning}, without access to global path information for obstacle avoidance, i.e., we exclude sub-goal knowledge from the observations.
\cref{fig:Baseline} presents the resulting learning curves. 
As expected, the baseline with path guidance converges. 
However, when this global guidance is removed, we observe convergence failure. 
This demonstrates that the method struggles with the guidance-free pushing task we consider, in addition to failing to incorporate the target object’s orientation. Incorporating orientation would require performing a 3D shortest‐path search over the full configuration space $(x,y,\theta)$, which either forces coarse discretization assumptions or incurs a prohibitively large search space and planning time that undermines online adaptability. Consequently, we do not pursue further experimental evaluation of this baseline under orientation‐aware criteria.

\begin{figure}[!t]
	\centering
	\includegraphics[width=.85\linewidth,  trim={0 0 0 0},clip]{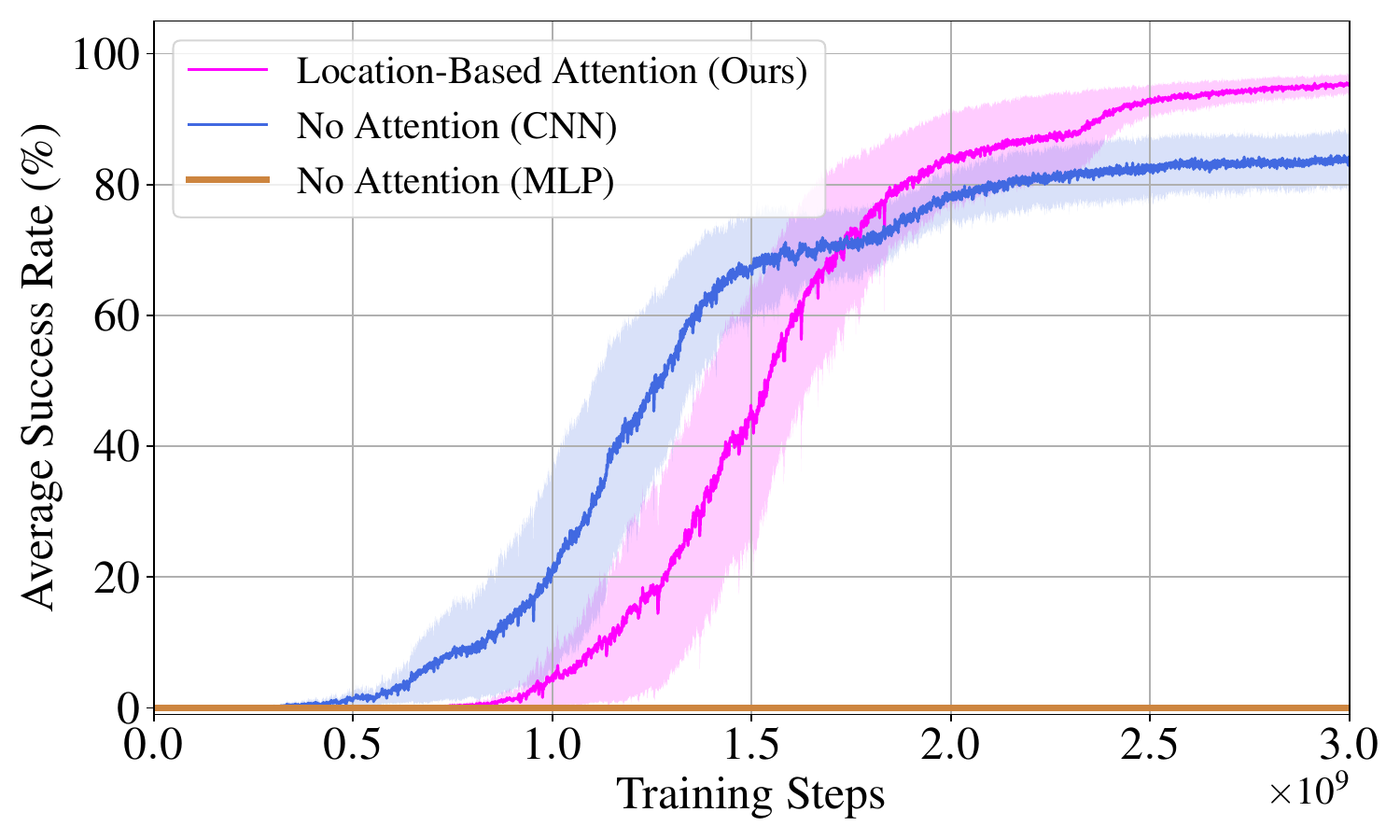}
	\caption{Training performance on our obstacle avoidance pushing task, with (Ours) and without (CNN) attention for feature extraction.} 
	\label{fig:training_convergence}
    \vspace{-5px}
\end{figure} 

\begin{figure*}[t] \centering 
\subfloat[Start configuration\label{fig:real_eval:a}]{\includegraphics[width=.175\textwidth, trim=0 100 0 75, clip]{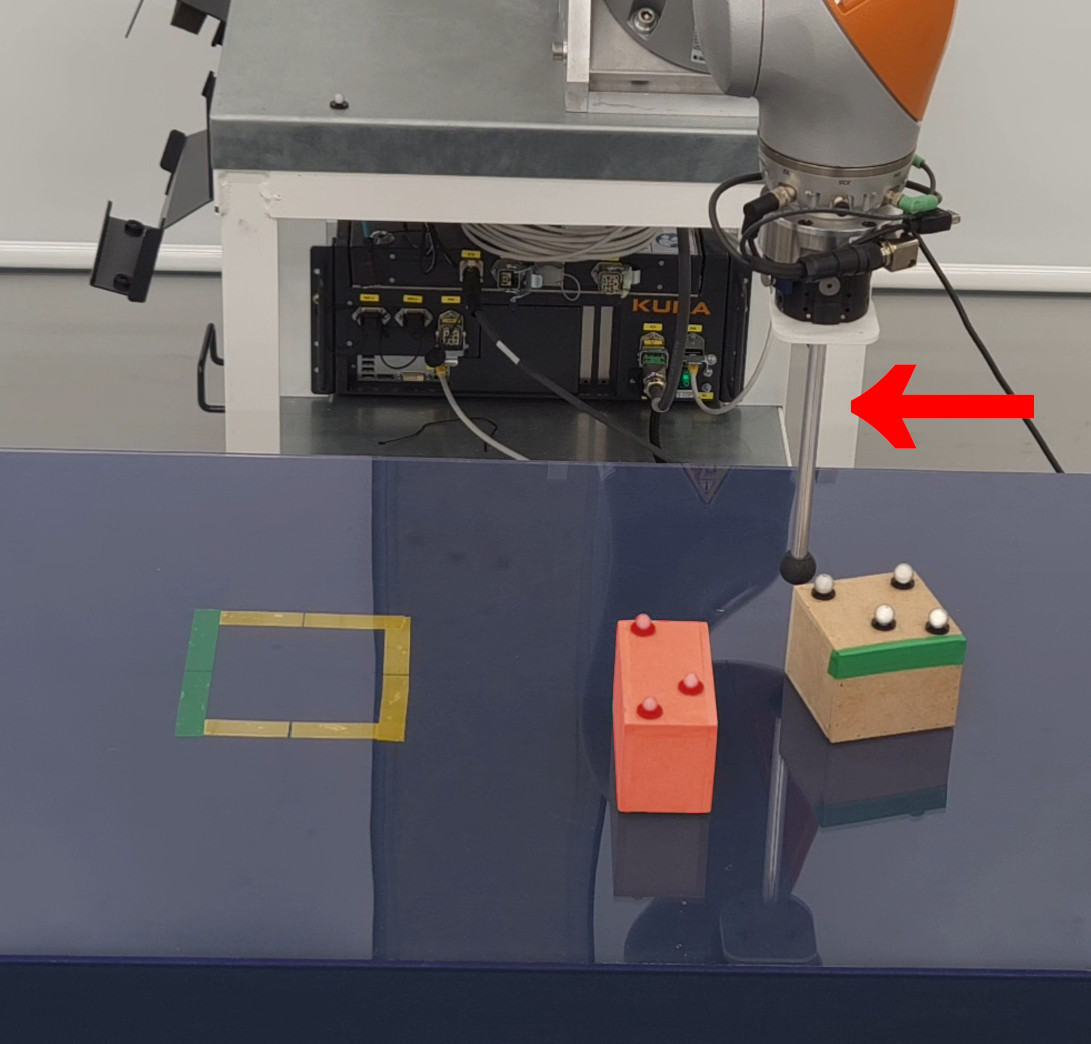}} \hspace{0.2em} 
\subfloat[\label{fig:real_eval:b}]{\includegraphics[width=.175\textwidth, trim=0 100 0 125, clip]{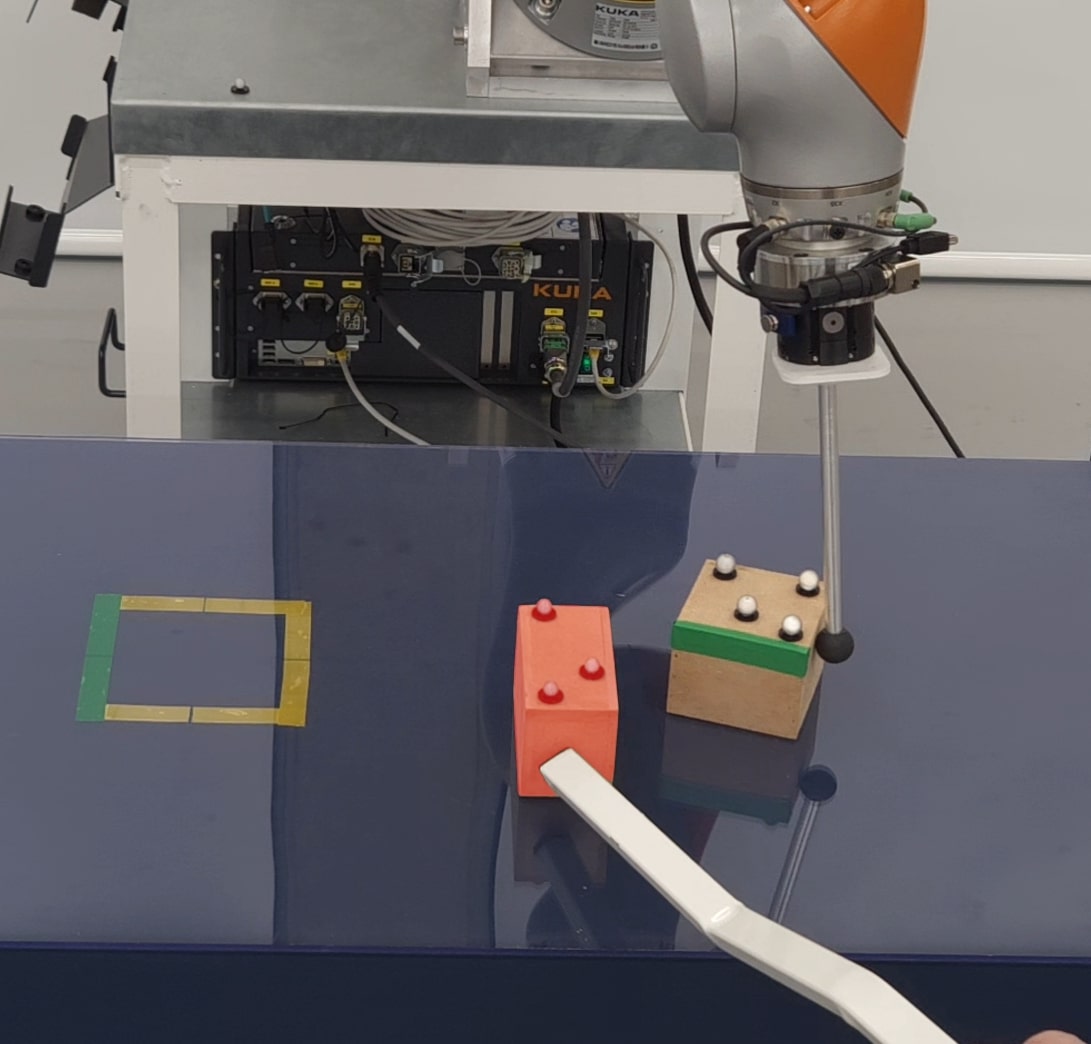}} \hspace{0.2em} 
\subfloat[\label{fig:real_eval:c}]{\includegraphics[width=.175\textwidth, trim=0 100 0 70, clip]{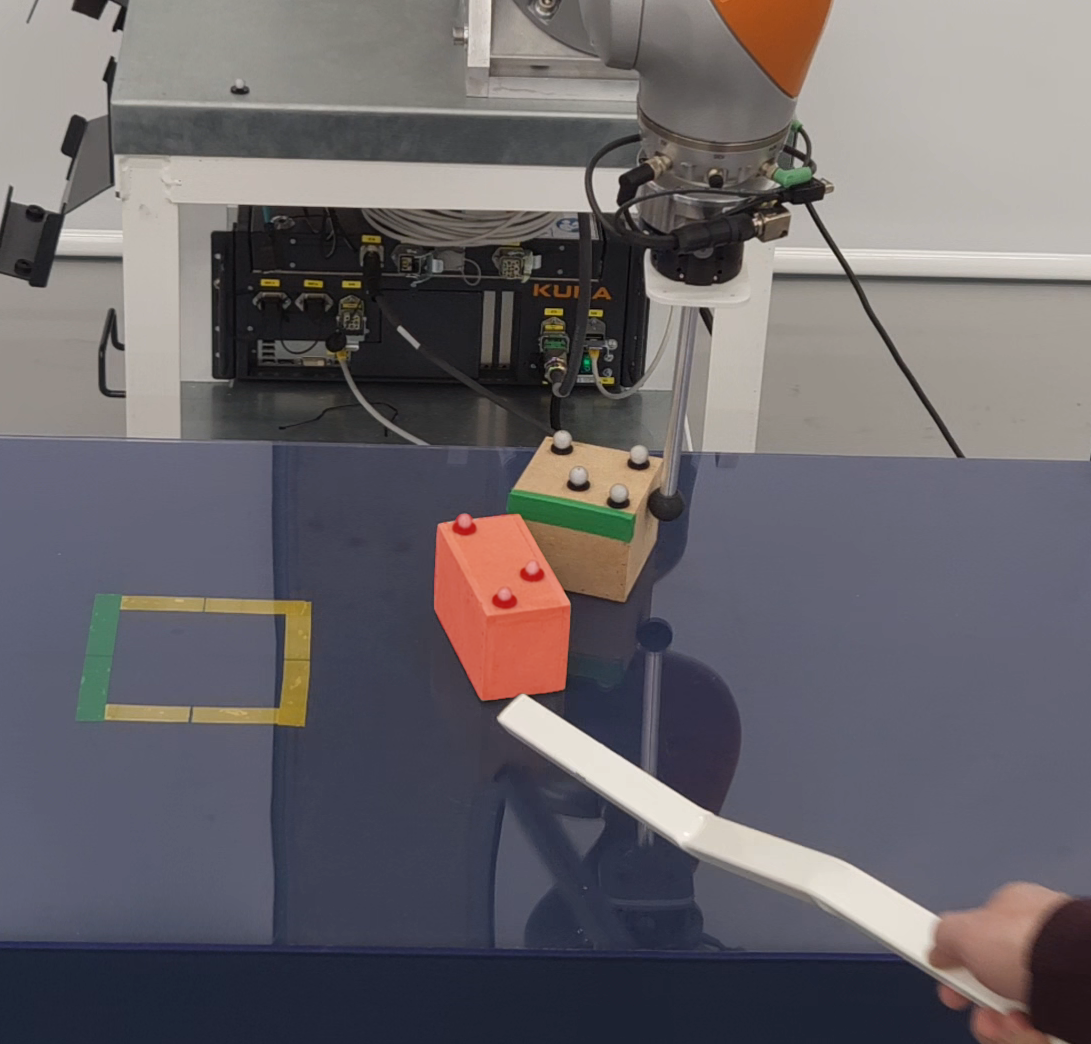}}  \hspace{0.2em} 
\subfloat[\label{fig:real_eval:d}]{\includegraphics[width=.175\textwidth, trim=0 100 0 70, clip]{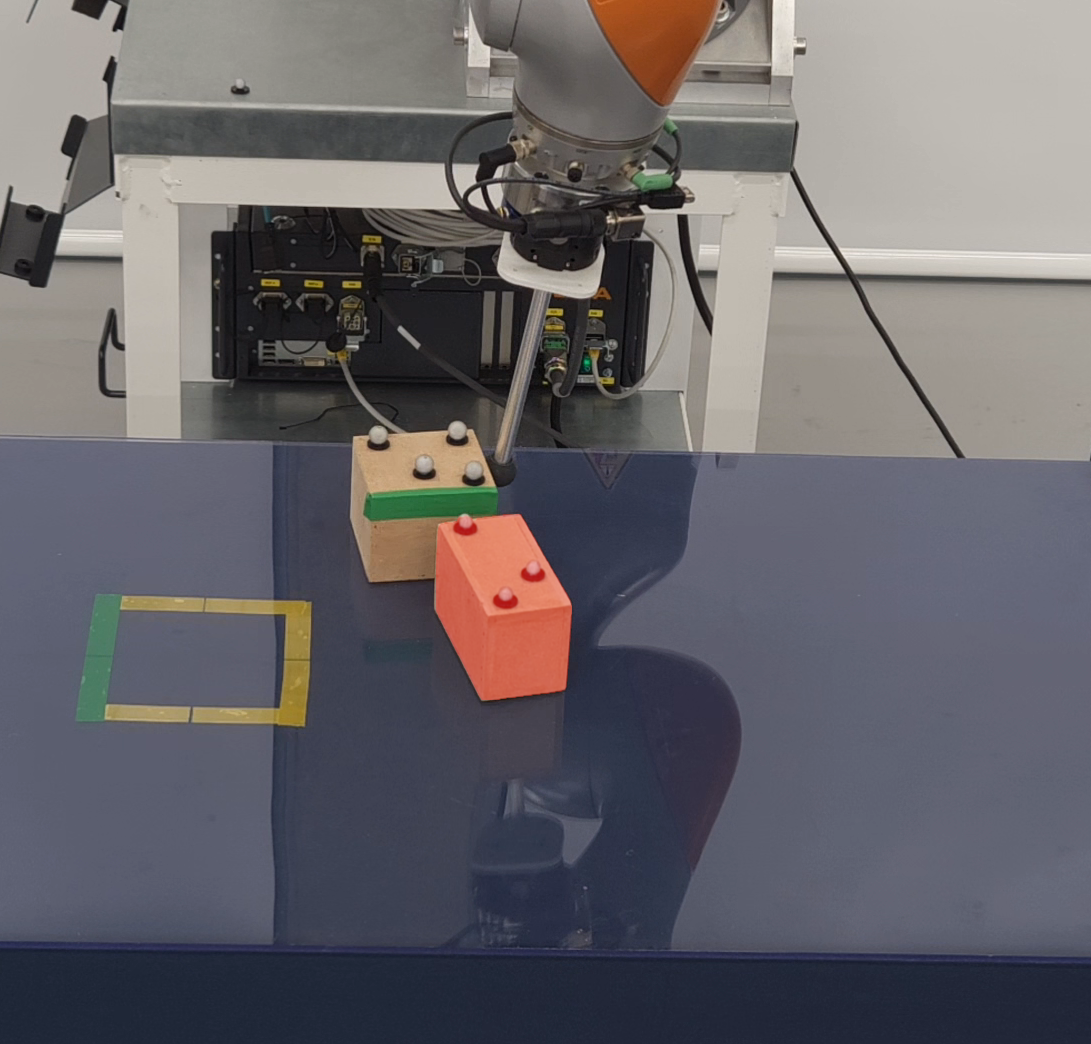}} \hspace{0.2em}
\subfloat[Target configuration\label{fig:real_eval:e}]{\includegraphics[width=.175\textwidth, trim=0 100 0 70, clip]{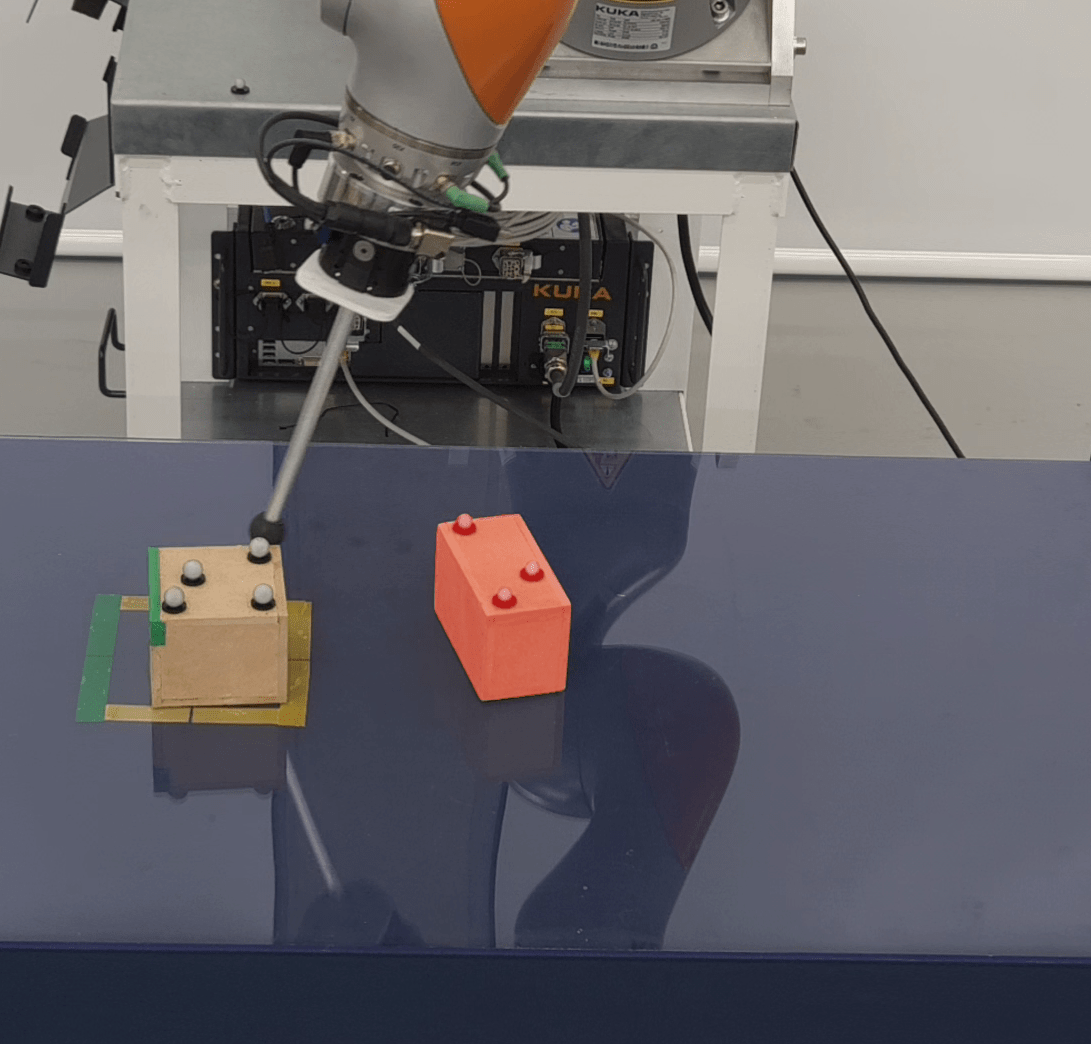}} 
\vspace{-2px}
\caption{
Key frames of the robot pushing an object from the start (a) to the target (e) configuration while avoiding a moving obstacle~(red).}
\label{fig:real_eval}
\end{figure*} 

\begin{figure*}[t] \centering 
\vspace{-18px}
\subfloat[Start configuration\label{fig:real_eval_cam:a}]{\includegraphics[width=.189\textwidth, trim=570 150 0 0, clip]{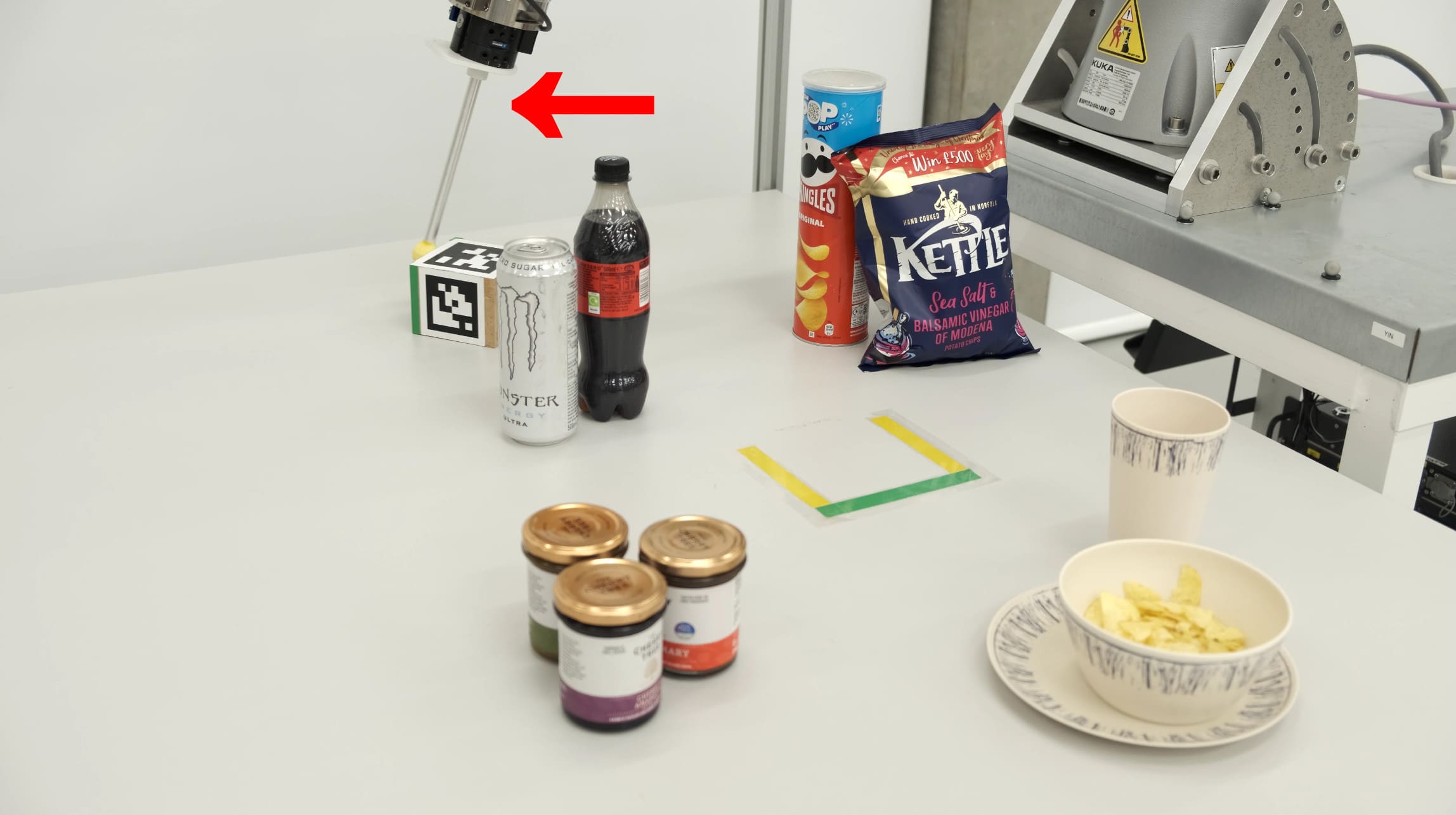}} \hspace{0.2em} 
\subfloat[\label{fig:real_eval_cam:b}]{\includegraphics[width=.189\textwidth, trim=570 150 0 0, clip]{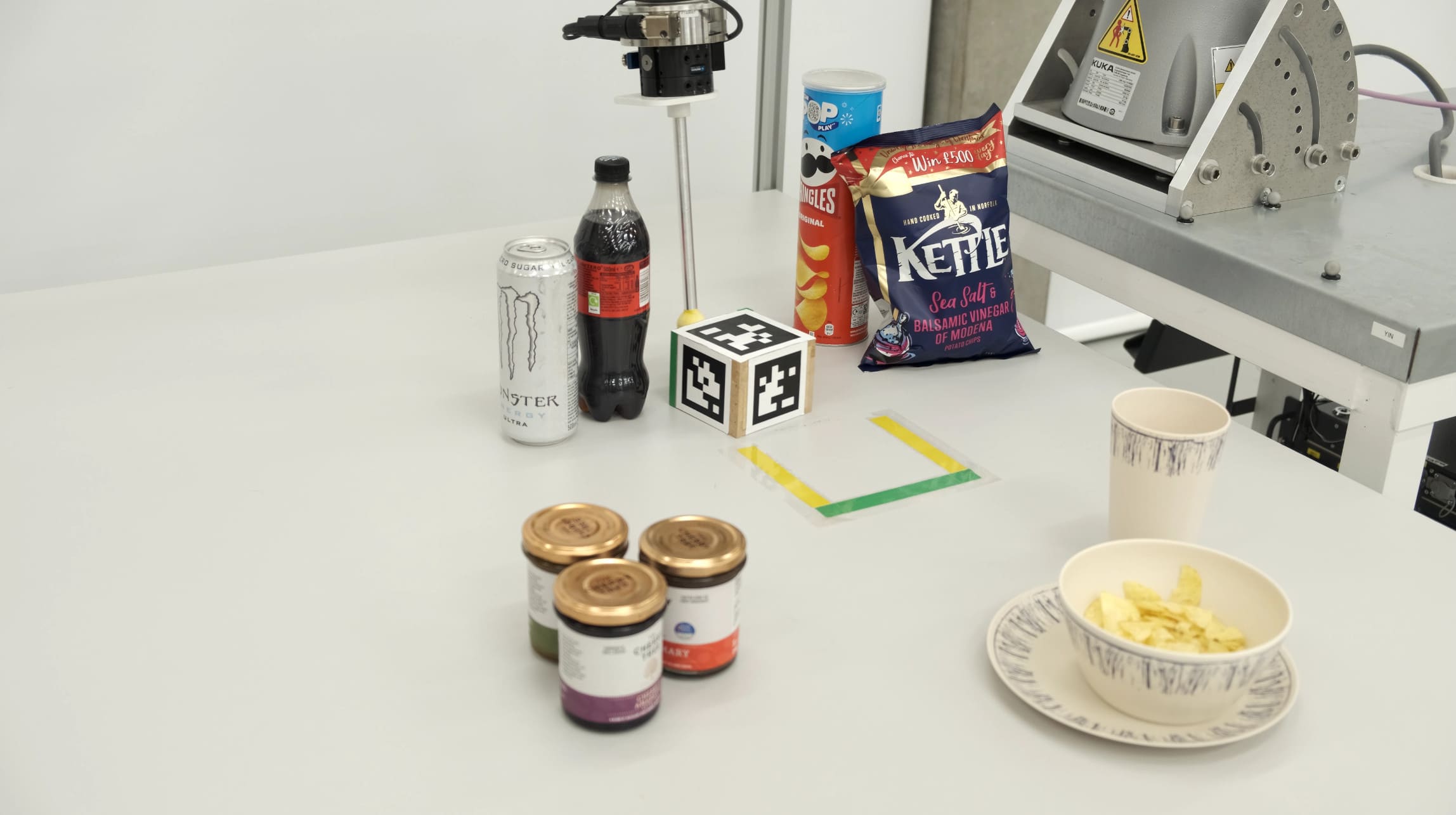}} \hspace{0.2em} 
\subfloat[\label{fig:real_eval_cam:c}]{\includegraphics[width=.189\textwidth, trim=0 150 0 0, clip]{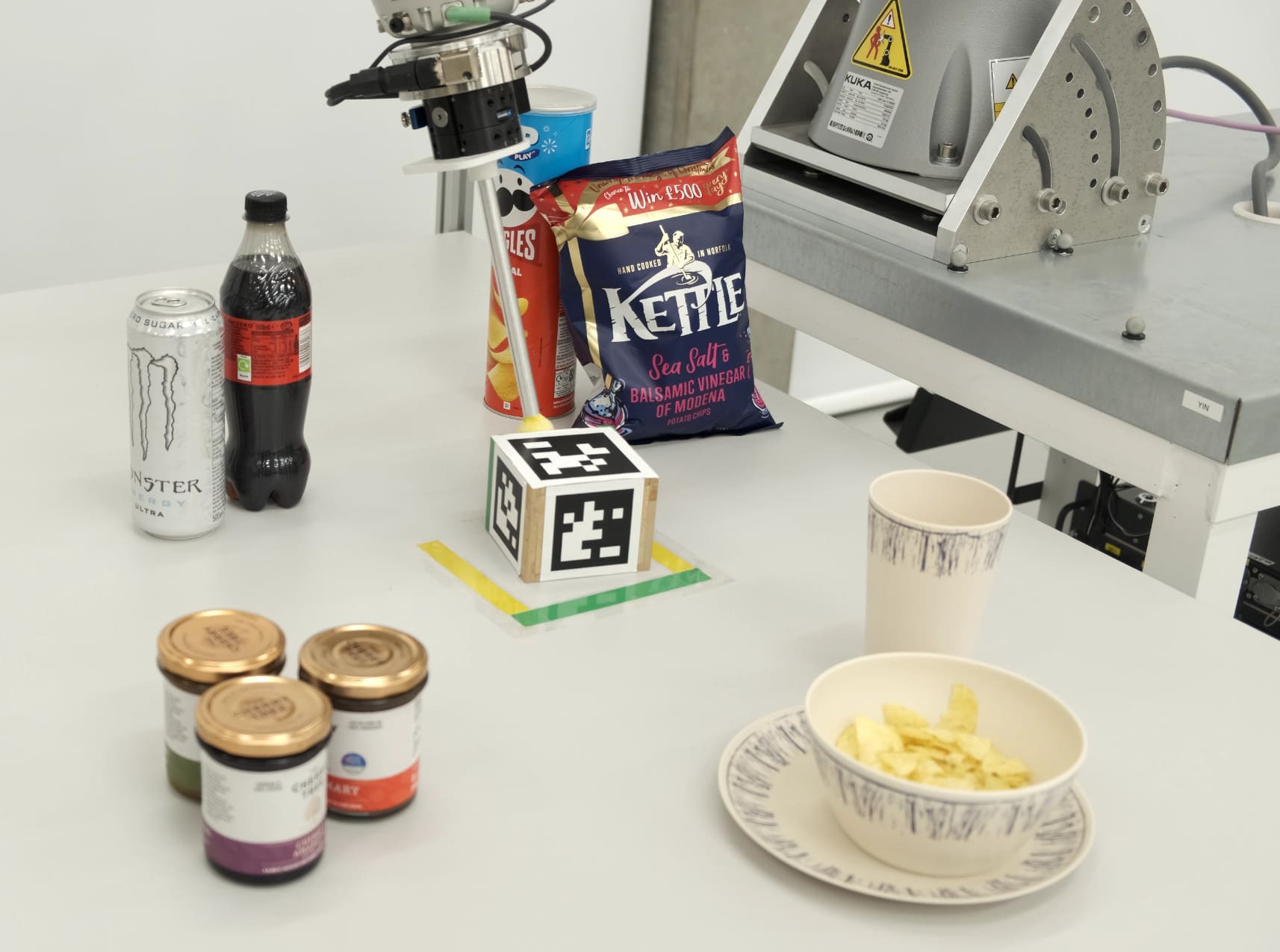}}\hspace{0.2em} 
\subfloat[\label{fig:real_eval_cam:d}]
{\includegraphics[width=.189\textwidth, trim=570 150 0 0, clip]{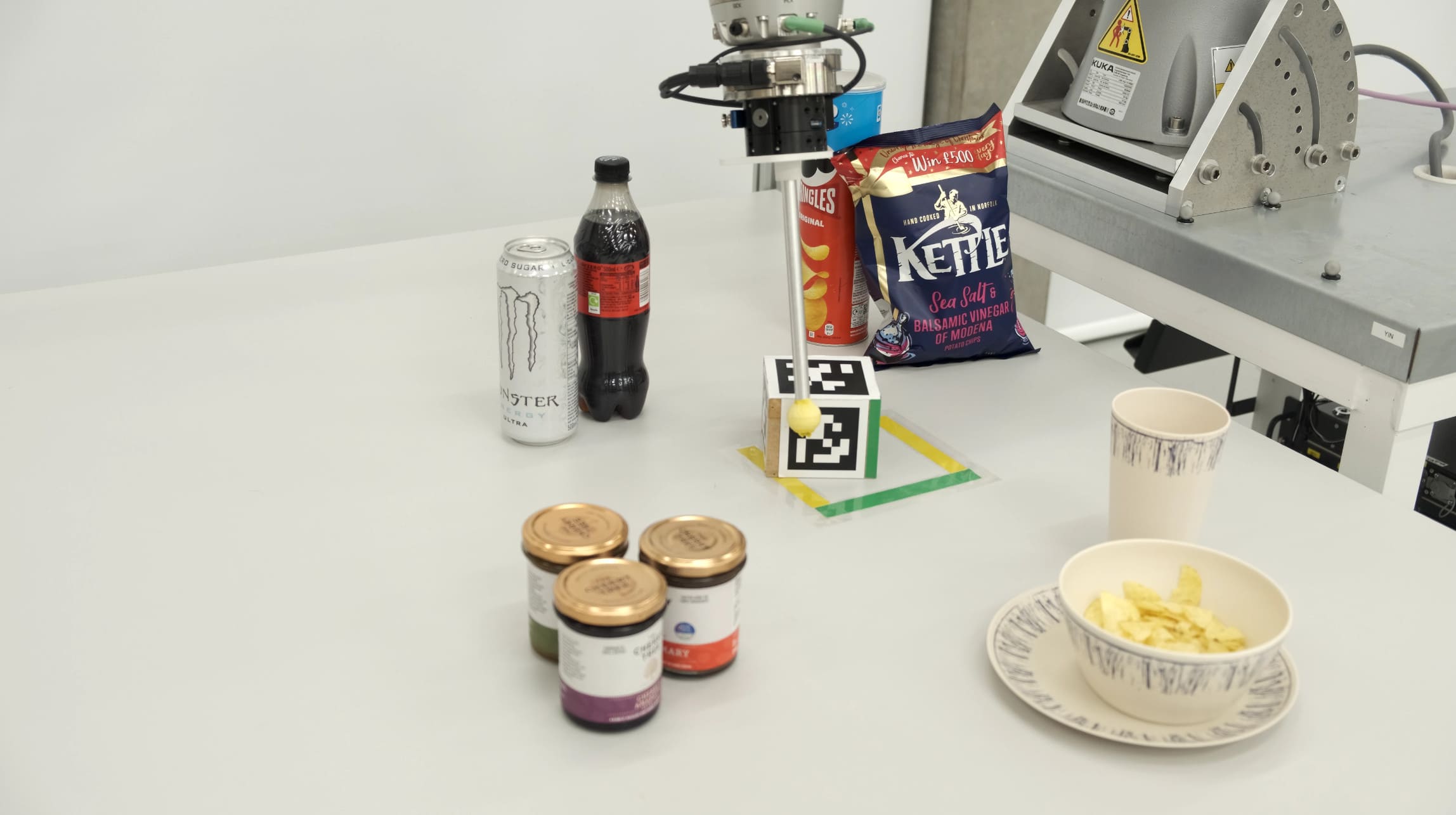}}  \hspace{0.2em} 
\subfloat[Target configuration\label{fig:real_eval_cam:e}]{\includegraphics[width=.189\textwidth, trim=570 150 0 0, clip]{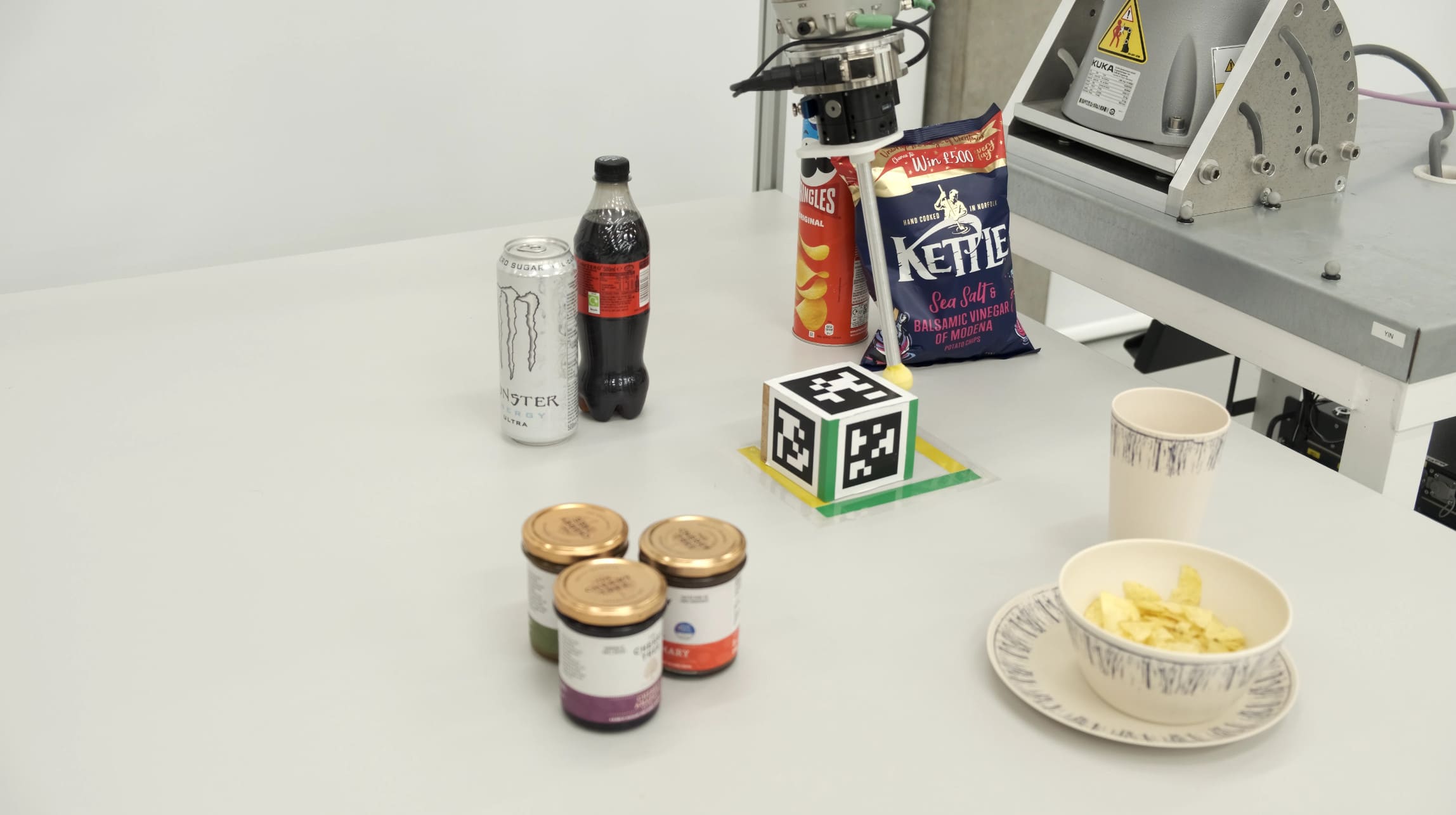}} \hspace{0.2em}  
\caption{Pushing an object from the start (a) to the target (e) configuration while avoiding multiple obstacles of different shapes.}
\vspace{-10px}
\label{fig:real_eval_cam}
\end{figure*} 

\subsection{Impact of Location-Based Attention on the Training}
\label{impact}

To investigate the impact of the location-based attention module, we compare it against alternative approaches for processing the occupancy grid during training and rollout.
In particular, we additionally implement a standard convolutional neural network (CNN) structure for feature extraction, using three CNN layers.
We also consider an ablation of our method that removes the computation of the weighted attention score sum, instead concatenating the feature vectors and compressing them through an MLP of size~\mbox{[2048, 512, 64]}.
Note that both alternative approaches have approximately the same number of learnable parameters as ours, ensuring a comparable model capacity.
Furthermore, we experimented with a multi-headed self-attention module (MHA), common in vision transformers, to compare it with our location-based attention approach. 
However, its high memory demands made training infeasible on an NVIDIA A6000 (48GB VRAM). 
The quadratic growth in memory usage, due to computing pairwise token interactions, severely limited our massively parallel simulations, causing a prohibitive slowdown in RL training. 
Given these constraints, we excluded MHA from the final training comparison.

\figref{fig:training_convergence} shows the resulting training curves for our proposed framework as well as the CNN and MLP modified approaches for processing the occupancy map.
We report mean and standard deviation across three training seeds.
We find that our approach with location-based attention achieves the highest final success rate ($96\%$).
On the other hand, while convergence is faster with the CNN structure, its asymptotic performance is noticeably lower ($87\%$). 
Furthermore, the CNN has a $70\%$ higher GPU memory consumption, due to the computational overhead from convolutional operations storing multiple large intermediate feature maps, making our method more efficient and with a better performance. 
Finally, the MLP ablation of our method, removing the weighted sum computation with attention scores, fails to converge, highlighting the critical role of selectively attending to spatial features.

\begin{table}[t]
    \centering
    \resizebox{\linewidth}{!}{
    \begin{tabular}{lcc|cc}
        \toprule
        \multirow{2}{*}{\textbf{Experimental Setup}} & \multicolumn{2}{c}{\textbf{Location Based Attention (Ours)}} & \multicolumn{2}{c}{\textbf{CNN Feature Extraction}} \\
        & \textbf{Success Rate \%} & \textbf{Collision Rate \%} & \textbf{Success Rate \%} & \textbf{Collision Rate \%} \\
        \midrule
        Training & 97.1 & 1.26 & 88.5 & 4.83 \\
        Circular & 95.6 & 2.66 & 84.7 & 0.56 \\
        Cross-Shape & 94.1 & 2.90 & 84.5 & 1.75 \\
        T-Shape & 93.5 & 4.72 & 85.3 & 0.97 \\
        L-Shape & 90.2 & 7.75 & 83.8 & 2.47 \\
        Dynamic & 84.3 & 12.0 & 73.5 & 14.9 \\
        \midrule
        Dual Obstacles & 48.1 & 50.7 & 57.9 & 34.3 \\
        Dual fine-tuned (DFT) & 91.2 & 3.54 & 61.1 & 3.22 \\
        Circular (DFT) & 96.4 & 0.20 & 72.1 & 0.34 \\
        Cross-Shape (DFT) & 96.7 & 0.33 & 73.8 & 0.54 \\
        T-Shape (DFT) & 96.3 & 1.32 & 71.9 & 1.01 \\
        L-Shape (DFT) & 94.9 & 1.58 & 71.2 & 1.22 \\
        Dynamic (DFT) & 91.4 & 5.2 & 61.6 & 10.8 \\
        \bottomrule
    \end{tabular}
    }
 \caption{Performance comparison between location-based attention (Ours) and CNN feature extraction for different obstacle configurations varying in size, shape, and quantity. 
    We report success and collision rates averaged across 2,000 randomized episodes.  
    Our method demonstrates significantly higher success rates across all scenarios and especially a superior fine-tuning capabiliy to novel scenes.
    The remaining failure cases beyond collisions are due to time out and workspace boundary violations.}
    \vspace{-5px}
    \label{tab:quantitative}
\end{table}
\vspace{-5px}
\subsection{Quantitative Evaluation}
 \label{sec:quantitative}
We conduct a quantitative evaluation of our framework and compare it against the baseline CNN feature extraction described in Sec. \ref{impact}. Our evaluation is performed 
across multiple environment configurations, incorporating various unseen obstacle shapes, and sizes. 
Specifically, we assess performance in environments containing circular, cross, T- and L-shaped obstacles, as well as a dual obstacle setup.  
Three of these configurations are illustrated in \figref{eval_environments}.
Furthermore, we evaluate on a dynamic-obstacle environment, where the obstacle moves along the y-axis at a constant speed of 0.1\si{\meter\per\second}, matching the pusher's maximum speed along the axis.
The obstacle reverses direction at the workspace boundaries. 
Note that the agent has never experienced a dynamic obstacle during training. 

We evaluate each trained policy for $2,000$ episodes per environment, with randomized start and target poses, as well as varying obstacle poses, sizes, and shapes. 
We consider an episode successful when the pusher and the manipulated object avoid collisions, the object remains within the workspace boundaries, it is placed within \SI{1.5}{\cm} and $\pi/6$ \SI{}{\radian} of the target pose, and the task completes in no more than 200~steps.

\cref{tab:quantitative} presents the results of this evaluation.
For single-obstacle scenarios, our method consistently achieves higher success rates, outperforming the CNN-based method across all tested obstacle shapes and the dynamic-obstacle environment. 
Although collision rates are slightly lower for the CNN baseline in the unseen shapes, this outcome largely stems from inaction---the policy often stops pushing completely---which in turn causes a significant increase in time-limit failures.
In contrast, even in scenarios with unseen obstacle shapes or unseen obstacle dynamics, our agent achieves high success rates and only rarely stops pushing, demonstrating its strong generalization capabilities.

We observe a notable performance gap in the dual obstacle scenario. 
When directly applying the single-obstacle-trained policies to this more challenging environment, our method achieves a $48.1\%$ success rate with a high collision rate of $50.7\%$, whereas the CNN-based approach performs better with $57.9\%$ success and $34.3\%$ collision rate. 
However, after fine-tuning on the dual obstacle environment~(DFT) for $5\cdot10^8$ steps, our method achieves a noticeably improved success rate of $91.2\%$ with a drastically reduced collision rate of $3.54\%$, demonstrating adaptability to more complex scenarios through targeted fine-tuning. 
In contrast, the CNN-based approach, even after fine-tuning, only reaches $61.1\%$ success with a $3.22\%$ collision rate, indicating limited adaptability to complex multi-obstacle environments.

Additionally, we evaluate the DFT agents on the single-obstacle environments and find that our method consistently improves success rates across all cases, despite being trained on a different dual obstacle scenario. 
Simply fine-tuning our agent on a more complex setting enables better generalization to both unseen shapes and unseen obstacle dynamics.
In contrast, the CNN baseline's success rate drops even further, highlighting its poor generalization and limited adaptability through fine-tuning.

These results suggest that while the CNN-based feature extraction provides a reasonable baseline, its feature representations are less effective, leading to reduced performance and severely limited adaptability through fine-tuning for highly constrained settings.
In contrast, our location-based attention method demonstrates superior adaptability and robustness, particularly when fine-tuned for more complex tasks. 
\vspace{-5px}
\subsection{Hardware Experiments}
\label{sec:real_world_exp}
For our physical hardware setup, shown in \cref{fig:cover}, we use a KUKA iiwa robot arm with OpTaS~\cite{Mower2023} to map the task-space policy actions to robot joint configurations.
To assess both precise action generation and real-world generalization, we explore two scene detection pipelines: a motion capture~(MoCap) system and an RGBD three-camera~(3Cam) setup. In the MoCap setup, a Vicon motion capture system tracks object and obstacle poses, directly generating the occupancy grid from it.
The 3Cam setup combines three Intel RealSense D435 cameras with AprilTags for object tracking and fuses point cloud data to construct the occupancy grid. 
While MoCap offers highly precise tracking and robustness against sensor noise, 3Cam provides greater flexibility for unstructured environments.
Note that for the hardware experiments, we fix the target pose to simplify the setup, but our simulation experiments fully randomize it.

To quantitatively evaluate our system's performance, we test 10 random initial configurations across three MoCap scenarios: (a) a standard setup with a single rectangular obstacle, (b) a single obstacle of an unseen shape, and (c) dual separated obstacles. 
\cref{eval_environments} shows smooth pushing sample trajectories generated by the physical robot in these scenarios.  
The learned policy achieves a $100\%$ success rate in (a) and (b), while in (c), it attains a $90\%$ success rate due to a single collision.

We also qualitatively evaluate the adaptability to dynamic obstacles. \cref{fig:real_eval} illustrates a scenario where the robot successfully pushes an object to the target pose while actively avoiding a moving obstacle. 
As the robot starts pushing, we dynamically reposition an obstacle to intersect the object's trajectory, significantly increasing task complexity. 

In the 3Cam setup, we test diverse obstacle configurations using everyday objects. 
\cref{fig:real_eval_cam} showcases a dining table scenario where the robot first pushes the object through a narrow gap and then re-orients it to reach the target pose, all while avoiding collisions. 
The supplemental video\footnote{\url{https://youtu.be/Ef0_oQiDq2E}\vspace{-.75em}} provides further demonstrations of our system’s performance in various real-world scenarios with both MoCap and 3Cam setups.
In all recorded scenarios, the robot exhibits smooth and continuous trajectory execution without relying on offline path planning.


\section{Conclusion}
\label{sec:conclusion}
In this paper, we present a model-free RL framework for non-prehensile planar pushing with obstacle avoidance in cluttered environments.
We leverage a computationally efficient location-based attention mechanism to extract and attend to relevant spatial features, and use categorical exploration during training to capture the multimodal nature of pushing. 
Unlike prior work, our framework removes the need for global path guidance and accounts for the manipulated object’s target orientation. 
By representing clutter with an occupancy grid, our system adapts to diverse and dynamically changing layouts.
Our experiments demonstrate that the learned policies achieve high success rates with low collision rates, even in configurations with unseen obstacle shapes, and can be efficiently fine-tuned for more complex scenarios involving multiple obstacles.
Finally, we evaluate the robustness of our approach in a physical hardware setup, demonstrating smooth and precise trajectories under various challenging clutter layouts, including dynamic obstacles.

%


\printbibliography

\end{document}